%% file: main.tex
\documentclass[11pt]{article}
\usepackage{acl}
\usepackage{times}
\usepackage{latexsym}
\usepackage[T1]{fontenc}
\usepackage[utf8]{inputenc}
\usepackage{microtype}
\usepackage{inconsolata}
\usepackage{graphicx}
\usepackage{amsmath}
\usepackage{stfloats}
\usepackage{booktabs}
\usepackage{makecell}

\title{A Dual-Track Framework for Template-Constrained LaTeX Conversion}
\author{
  Cheuk Hei Chung\thanks{chchung634@connect.hkust-gz.edu.cn} \quad Li Liu \thanks{Correspondance to: Li Liu <avrillliu@hkust-gz.edu.cn>} \\ 
  The Hong Kong University of Science and Technology (Guangzhou) \\ 
  }
\date{}

\usepackage{amsfonts}
\usepackage{amssymb}
\usepackage{booktabs}
\usepackage{natbib}
\begin{document}
\maketitle
\begin{figure*}[!htbp]
  \includegraphics[width=\linewidth]{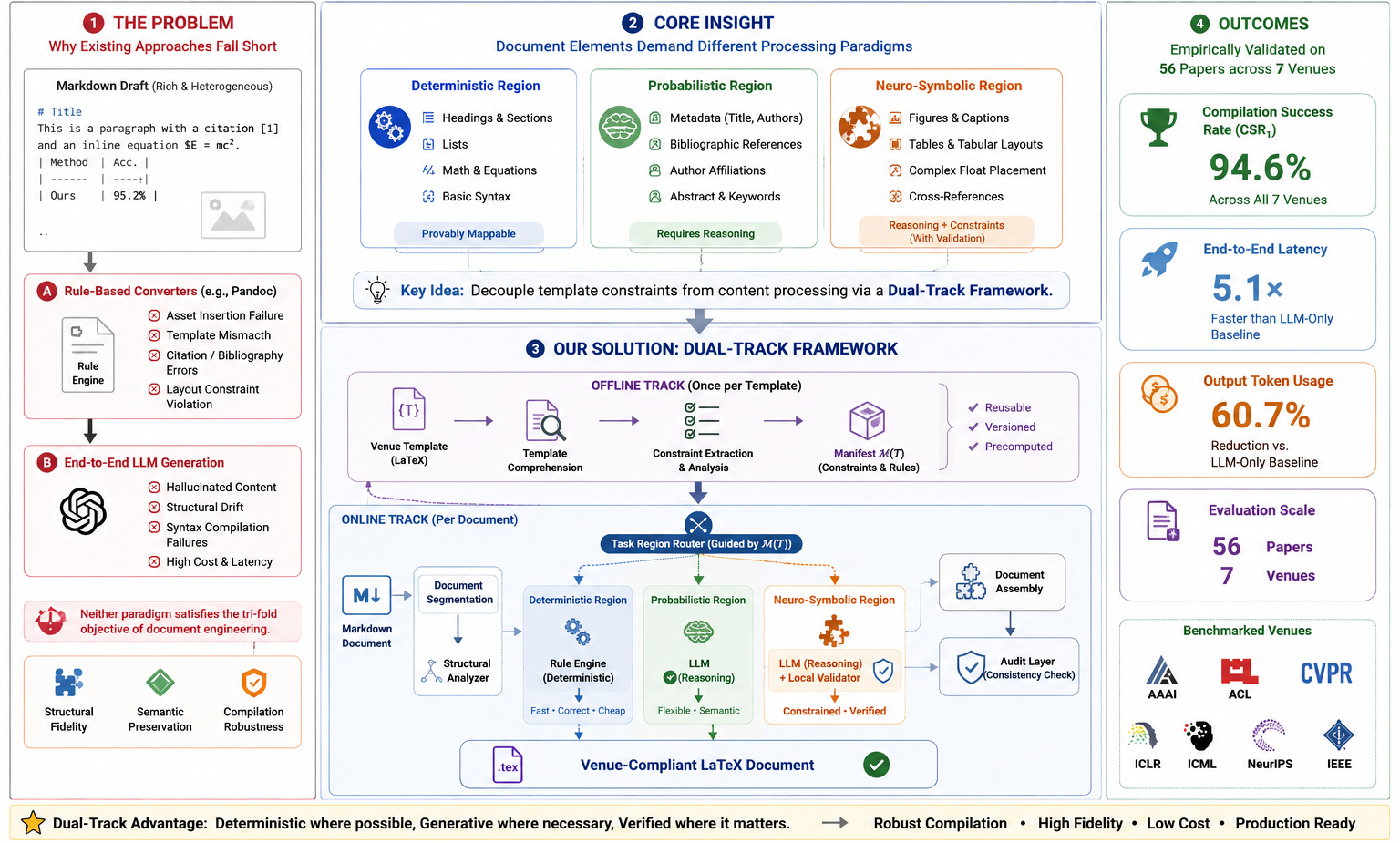}
  \caption{\textbf{Overview of our Dual-Track Framework}. We first analyze documents into three task regions—Deterministic, Probabilistic, and Neuro-Symbolic—each requiring a distinct processing paradigm. Our framework decouples template constraints (offline track) from document processing (online track), routing each region to the most suitable engine under the guidance of a precomputed manifest. Extensive evaluations demonstrate superior compilation robustness, efficiency, and semantic fidelity across diverse real-world templates and documents.}
  \label{fig:dual_track}
\end{figure*}

\begin{abstract}

\input{01_Abstract}

\end{abstract}

\input{02_Introduction}

\input{03_Task_Decomposition_and_System_Architecture}

\input{04_Experiments}

\input{05_Discussion}

\input{06_Conclusion}

\input{07_Limitations}

\input{08_Ethical_Considerations}

\bibliography{09_References}

\appendix
\input{10_Appendix_A}

\input{11_Appendix_B}

\input{12_Appendix_C}

\input{13_Appendix_D}

\input{14_Appendix_E}

\input{15_Appendix_F}
\input{16_Appendix_G}

\end{document}

%% file: 01_Abstract.tex
With the increasing demands for advanced document conversion, mapping structured Markdown drafts into template-compliant formats like LaTeX remains a challenge. Existing approaches largely depend on either deterministic rule-based converters or pure end-to-end Large Language Model (LLM) generation. The former fails to correctly handle asset insertions and template-specific constraints, while the latter tends to induce semantic drift, leading to hallucinations that are difficult to debug. To address these limitations, we introduce a robust \textbf{Dual-Track Framework} that systematically decouples template formatting from document processing: an offline track extracts template constraints into a reusable manifest, while an online track implements a \textbf{hybrid execution pipeline.} This pipeline confines LLM usage exclusively to reasoning-intensive components (e.g., semantic metadata, bibliographic references, and complex visual/tabular layouts) while delegating rule-based engines for deterministic processing. Empirical evaluation across 7 LaTeX templates and 56 published research papers demonstrates that our method preserves better structural fidelity, satisfies diverse layout constraints, and achieves a higher compilation success rate compared to the previous baselines.

%% file: 02_Introduction.tex
\section{Introduction}

Bridging structural gaps between Markdown drafts and venue-specific LaTeX templates remains a high-overhead problem. An ideal converter must achieve a tri-fold objective: \textbf{preserving source semantics}, \textbf{satisfying layout constraints}, and \textbf{ensuring robust compilation}. Existing methods fail to meet these concurrent requirements \citep{Li:2025jd}. Rule-based engines like Pandoc \citep{MacFarlane:2009qr} lack contextual reasoning for template nuances, while end-to-end Large Language Model (LLM) generation introduces structural hallucinations and fatal syntax errors that are difficult to audit \citep{Ji:2023qr,Liu:2024jd,Boudourides:2026qr,Dou:2026jd}.

This failure stems from a fundamental conflict: standardized syntactic structures are deterministic, making LLM processing expensive and error-prone where rule-based mapping is provably correct. Conversely, semantic metadata, citations, and complex layouts require linguistic reasoning that defies hard-coded rules \cite{Liu:2026jk}. Aligning with neurosymbolic design principles \citep{Roberts:2024jd,Garcez:2023jd}, document conversion naturally demands a hybrid paradigm.

Given these complementary properties, we present the \textbf{Dual-Track Framework}, a document transformation method that systematically decouples template formatting from text processing. The framework operates through an offline track that extracts template-specific formatting constraints into a reusable manifest and an online track that executes a hybrid pipeline. Adhering to a neurosymbolic paradigm \cite{Garcez:2023jd,Manhaeve:2018jd,Kautz:2022jd}, this pipeline routes deterministic markdown syntax to rule-based engines, while restricting LLM invocation strictly to reasoning-intensive components guided by the pre-computed manifest.

Evaluated on 56 published papers across 7 academic templates, our framework outperforms both rule-based baselines and monolithic generative pipelines \cite{Ouyang:2025jd,Li:2025jd,Lyn:2025jd}. By bounding generative behavior, it mitigates operational latency and token costs, providing a stable, production-ready solution. Our core contributions are:

\begin{itemize}
  \item We introduce the \textbf{Dual-Track Framework} that systematically decouples structural template compliance from content semantic translation, which significantly reduces runtime computational overhead.
  \item A hybrid execution pipeline is implemented within the framework's online track which confines LLM invocations strictly to reasoning-intensive components (semantic metadata, bibliographic references, and visual/tabular layouts) while delegating deterministic syntax to rule-based engines.
  \item Systematic evaluations and ablation studies across 7 major publishing venues using 56 published papers are provided. The results empirically validate that our framework successfully satisfies the tri-fold objective of document engineering by simultaneously preserving structural fidelity, semantic preservation and adhering to strict layout constraints.
\end{itemize}

%% file: 03_Task_Decomposition_and_System_Architecture.tex
\section{Task Decomposition and System Architecture}

This section formalizes the template-constrained document conversion problem and introduces the architecture and deployment of our Dual-Track framework that decomposes document processing based on structural predictability.

\subsection{Problem Characteristics and Formulation}

\textbf{Problem Formulation.} The objective of template-constrained document conversion is to transform a source Markdown document $D_{\text{md}}$ into a compilable LaTeX artifact $D_{\text{tex}}$ that strictly complies with a target venue template $\mathcal{T}$ (comprising class files, style configurations, and layouts). The output $D_{\text{tex}}$ must simultaneously satisfy a tri-fold objective:

\begin{itemize}
  \item \textbf{Structural Fidelity:} $D_{\text{tex}}$ must strictly conform to the template-specific syntactic invariants dictated by $\mathcal{T}$ (e.g., preamble declarations, front-matter layout, and bibliography modes).
  \item \textbf{Semantic Preservation:} Accurate translation and preservation of all textual and asset elements from $D_{\text{md}}$ without historical data loss.
  \item \textbf{Compilation Robustness:} Successful compilation under the prescribed LaTeX engine without fatal errors.
\end{itemize}
This formulation addresses a core engineering conflict: Markdown possesses a flat, loosely constrained syntax, whereas LaTeX is a nested, macro-driven language bound by strict, template-specific invariants that preclude monolithic generative or inflexible rule-based pipelines \cite{Li:2025jd, Lyn:2025jd}. A  mathematical formalization of this dual-track mapping is provided in Appendix A.

\textbf{Task Region Taxonomy.} To optimize both execution cost and processing robustness, the system processes a source document by routing its elements into three functional regions:

\begin{itemize}
  \item \textbf{Deterministic Region:} Elements where syntax mapping is entirely predictable (e.g., basic syntax mapping) and are routed directly to a high-efficiency rule-based engine.
  \item \textbf{Probabilistic Region:} Elements requiring open-ended linguistic or contextual reasoning (e.g., front-matter metadata alignment and bibliographic references) which are delegated to LLM inference guided by the pre-computed template manifest.
  \item \textbf{Neuro-Symbolic Region:} Elements requiring generative neural processing coupled with strict syntax constraints (e.g., text chunks embedded with tables/figures, or complex asset layouts) where LLM-generated snippets are immediately intercepted and validated by a localized deterministic validation layer before final serialization \cite{Garcez:2023jd, kwon2025neuro, chatzikyriakidis2026neuro}.
\end{itemize}

By executing this taxonomy, the framework restricts LLM deployment exclusively to regions where rule-based transformation is intractable, securing deterministic reliability for standard syntax while leveraging neural reasoning where necessary.

\subsection{System Architecture}

The Dual-Track Framework decouples offline template comprehension from online document execution through six sequential stages ($T_1$ to $T_6$), as outlined in Table \ref{tab:arch_overview}.

\begin{table*}
  \centering
  \small
  \begin{tabular}{|l|l|l|l|l|}
    \hline
    \textbf{Stage} & \textbf{Track} & \textbf{Operation Name} & \textbf{Workload Classification} & \textbf{Primary Mechanism} \\
    \hline
    $T_1$ & Offline & Template Comprehension & Neuro-Symbolic & LLM Parsing \& Rule Verification \\
    \hline
    $T_2$ & Online & Document Segmentation & Deterministic & Regex \& Markdown AST Parser \\
    \hline
    $T_3$ & Online & Hybrid Chunk Routing & Neuro-Symbolic & Asset targeted LLM \& Syntax mapping \\
    \hline
    $T_4$ & Online & Inline Structural verification & Deterministic & Rule-Based Validation \& Auto-Repair \\
    \hline
    $T_5$ & Online & Global Document Assembly & Probabilistic & Front-matter \& BibTeX Generation \\
    \hline
    $T_6$ & Online & Post-Processing \& Audit & Deterministic & Citation Resolution \& Logging \\
    \hline
  \end{tabular}
  \caption{Architectural overview of the six sequential stages across the offline and online tracks.}
  \label{tab:arch_overview}
\end{table*}.

\subsubsection{Offline Track: Template Comprehension}

Independent from document processing, an LLM evaluates the target template files $\mathcal{T}$ to extract formatting constraints and package requirements. These properties are serialized into a reusable Manifest $\mathcal{M}(\mathcal{T})$, eliminating template-reasoning overhead from the runtime path.

\subsubsection{Online Track: Hybrid Execution Pipeline}

For each document $D_{\text{md}}$, the online track ingests Manifest $\mathcal{M}(\mathcal{T})$ as a static guide and processes the data through the hybrid pipeline ($T_2$ to $T_6$).

\begin{itemize}
  \item \textbf{Syntax Routing and LLM Isolation ($T_2, T_3$):} The framework uses a rule-based parser to segment the document. Non-asset chunks bypass neural inference entirely and are processed via deterministic syntax mapping. LLM invocation is strictly bounded to local coordinates containing complex asset placeholders or dense semantic contexts \citep{Liu:2024jd}.
  \item \textbf{Localized Structural Validation ($T_4$):} To prevent localized generative hallucinations from corrupting document structure, an inline validation layer intercepts the LLM-generated segments from $T_3$ and applies automated rule-based patches to fix basic errors prior to assembly \cite{poesia2022synchromesh}.
  
  \item \textbf{Global Assembly and Compilation Audit ($T_5, T_6$):} The pipeline merges the validated segments, leverages the LLM for front-matter metadata command alignment and generates the bibliography file. In the final stage, a deterministic engine resolves in-text citation and automatically isolates unresolvable citation keys into an external audit file for manual review, ensuring end-to-end compilability of the generated bundle.
\end{itemize}

\subsection{Deployment}
The framework is deployed as a production web service utilizing a self-contained architecture to minimize infrastructure complexity \cite{fehlis2026operationalizing}. To support heavy document workflows, job scheduling and state management rely on database-backed persistence rather than external messaging queues. Document submissions advances through a linear state machine aligned with stages $T_2$ to $T_6$, providing crash recovery from worker interruptions or service restarts \cite{gosmar2026madp,chen2026stratus}. To maximize the online throughput of the parallel execution layer $T_3$, pre-computed template manifests $\mathcal{M}(\mathcal{T})$ are distributed as versioned build artifacts during service deployment \cite{xu2025ragops}. The deployment layer also strengthens API integrations against network instability and timeouts using isolated connection pools and exponential backoff strategies. Finally, our service tracks each job with lightweight, end-to-end trace IDs, allowing operators to cross-reference compilation failures and monitor the parsing lifecycle without a dedicated observability stack.

%% file: 04_Experiments.tex
\section{Experiments}

To evaluate our framework under a reproducible industrial setting, we constructed a benchmark dataset of 56 research papers totaling over 900 pages spanning 7 major venues (AAAI, ACL, CVPR, ICML, ICLR, NeurIPS, IEEE), excluding surveys and workshop papers. We evaluate against rule-based Pandoc and a monolithic E2E LLM pipeline. Both the E2E baseline and our framework utilize OpenAI-compatible Qwen-Plus \citep{yang2024qwen2technicalreport} API, selected for its balanced cost-quality ratio, extended context window, and stable API responsiveness.

\subsection{Experimental Setup}

We employ a \textbf{Dual-Anchor Round-Trip Evaluation Framework}, inspired by round-trip evaluation paradigms in structure extraction tasks \citep{Roberts:2024jd}, to quantify performance against the tri-fold objective: structural fidelity, semantic preservation, and compilation robustness. A \textbf{LaTeX Ground-Truth($D_{\text{tex}}^{\text{gt}}$)} is derived from the main body of a source package into a single root LaTeX file to be the standard for template compliance. $D_{\text{tex}}^{\text{gt}}$ is then converted into an initial Markdown stream and sanitized to produce the clean \textbf{Markdown Ground Truth ($D_{\text{md}}$)}, which closely mirrors a human-authored draft (with the \textit{References} section removed to enforce dynamic citation resolution) to measure content faithfulness \citep{Li:2025jd}.

To ensure informational fairness, each pipeline receives identical Markdown Ground Truth ($D_{\text{md}}$) and raw \texttt{.bib} files. Our framework utilizes the venue-specific Manifest $\mathcal{M}(\mathcal{T})$ pre-computed by the offline track while baselines consume a standard \texttt{scaffold.tex} skeleton. Details regarding dataset selection and preparation are provided in Appendix B. This enables outputs to be systematically cross-examined against the dual ground-truth anchors ($D_{\text{tex}}^{\text{gt}}$ and $D_{\text{md}}$) and across two distinct evaluation scopes:

\begin{itemize}
  \item \textbf{Primary Track (Structural \& Layout):} ,This track measures template compliance, compilation robustness, and execution efficiency through a local TeXLive 2026 compilation pass executed in robust mode. Compilation reliability is quantified with \textbf{One-Shot Compilation Success Rate ($\text{CSR}_1$)} and Average Fatal Errors ($\bar{e}$). To evaluate structural preservation, we use structural retention rates, specifically Sectioning commands ($\text{SRR}$), Floating environments ($\text{FRR}$), and Math environments ($\text{MathER}$), cross-examined against both $D_{\text{tex}}^{\text{gt}}$ and $D_{\text{md}}$ \citep{Kale:2025jd}. Efficiency metrics include recording total token consumption and operational wall-clock latency.
  \item \textbf{Supplementary Track (Semantic \& Citation)}: Conducted exclusively against the $D_{\text{tex}}^{\text{gt}}$, this track examines linguistic fidelity with ROUGE-L \citep{Lin:2004jd} and BERTScore F1 \citep{Zhang:2019jd} and citation metrics including Key Mapping Accuracy and Form Accuracy (e.g., \texttt{\textbackslash{}citet} vs. \texttt{\textbackslash{}citep}) verified by BibTeX parsing.
\end{itemize}

\subsection{Quantitative Results and Analysis}

\subsubsection{Primary Track Evaluation}

\begin{figure*}[!htbp]
  \includegraphics[width=\linewidth]{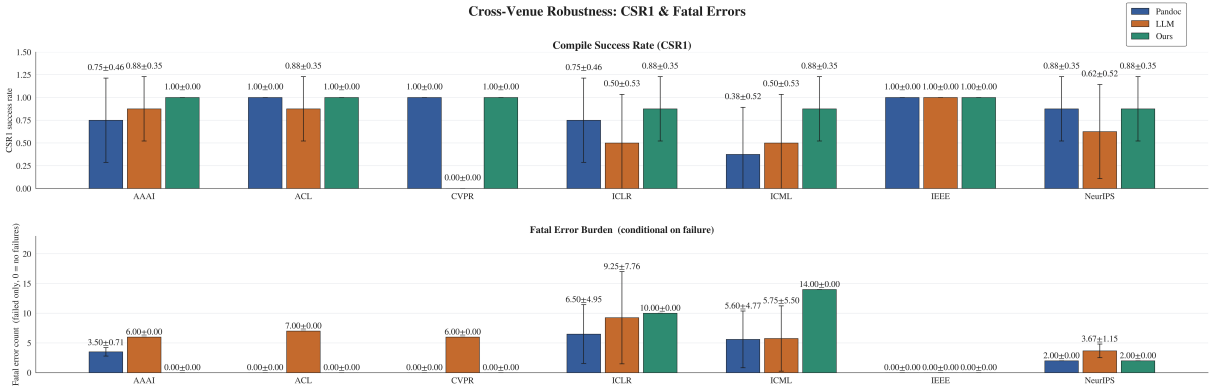}
  \caption{Bar Chart illustrating Cross Venue Robustness}
  \label{fig:cross_venue_robustness}
\end{figure*}

Table \ref{tab:aggregate_metrics} summarizes the aggregate quantitative metrics while the specific cross-venue distribution for compilation success is illustrated in Figure \ref{fig:cross_venue_robustness}.

\begin{table*}[t]
\centering
\makebox[\textwidth][c]{%
\footnotesize
\begin{tabular}{lcccccccccccc}
\toprule
Method &
CSR$_1$ &
SRR$_{\text{tex}}$ &
FRR$_{\text{tex}}$ &
MathER$_{\text{tex}}$ &
SRR$_{\text{md}}$ &
FRR$_{\text{md}}$ &
MathER$_{\text{md}}$ &
Fatal Avg &
Fatal &
Wall(s) &
InTok &
OutTok \\
\midrule
Pandoc &
\makecell{0.821 \\ \small $\pm$0.386} &
\makecell{1.028 \\ \small $\pm$0.091} &
\makecell{1.173 \\ \small $\pm$0.742} &
\makecell{0.895 \\ \small $\pm$0.200} &
\makecell{1.032 \\ \small $\pm$0.018} &
\makecell{0.863 \\ \small $\pm$0.225} &
\makecell{0.999 \\ \small $\pm$0.015} &
\makecell{5.0 \\ \small $\pm$3.9} &
50 &
0.1 &
0 &
0 \\
LLM &
\makecell{0.625 \\ \small $\pm$0.489} &
\makecell{0.988 \\ \small $\pm$0.110} &
\makecell{1.022 \\ \small $\pm$0.512} &
\makecell{1.331 \\ \small $\pm$0.598} &
\makecell{0.991 \\ \small $\pm$0.065} &
\makecell{0.881 \\ \small $\pm$0.511} &
\makecell{1.463 \\ \small $\pm$0.527} &
\makecell{6.3 \\ \small $\pm$4.1} &
132 &
208.5 &
25780 &
13927 \\
\textbf{Ours} &
\makecell{\textbf{0.946} \\ \small $\pm$0.227} &
\makecell{0.970 \\ \small $\pm$0.089} &
\makecell{1.432 \\ \small $\pm$0.748} &
\makecell{0.896 \\ \small $\pm$0.201} &
\makecell{0.972 \\ \small $\pm$0.036} &
\makecell{1.129 \\ \small $\pm$0.328} &
\makecell{\textbf{1.000} \\ \small $\pm$0.001} &
\makecell{8.7 \\ \small $\pm$6.1} &
\textbf{26} &
\textbf{40.6} &
\textbf{19550} &
\textbf{5477} \\
\bottomrule
\end{tabular}}
\caption{Aggregate evaluation metrics on the primary track (where $SRR$, $FRR$, and $MathER$ metrics are symmetrically cross-examined against both $D_{\text{tex}}^{\text{gt}}$ and $D_{\text{md}}$ anchors).}
\label{tab:aggregate_metrics}
\end{table*}

\textbf{Compilability \& Robustness.}
The aggregate metrics for the Primary Track are summarized in Table~\ref{tab:aggregate_metrics}. Our framework demonstrates superior robustness, achieving a $CSR_1$ of $0.946$ compared to $0.821$ (Pandoc) and $0.625$ (LLM). Despite a higher per-instance failure error count ($8.7$) due to a smaller sample of cases, our framework substantially reduces Total Fatal Errors to $26$ (vs. $132$ for LLM), indicating a lower frequency of systemic failures. Figure 1 illustrates this stability across venues; while the monolithic LLM collapses in complex environments (e.g., $CSR_1=0$ in CVPR), our framework maintains consistent compilation success rates. Failure Analysis for Primary Track are provided in Appendix F.

\textbf{Narrative \& Mathematical Fidelity.} Sectioning retention ($SRR$) remains stable across all paradigms. In contrast, mathematical environments highlight generative drift: the LLM baseline inflates $MathER$ to $1.331$ ($D_{\text{tex}}^{\text{gt}}$) and $1.463$ ($D_{\text{md}}$). Our framework stays closely aligned with the Pandoc baseline and preserves display math faithfully ($MathER_{\text{tex}}$=$0.896$ and $MathER_{\text{md}}$=$1.000$) , proving that the hybrid pipeline helps mitigate typical generative drift.

\textbf{Float Asset Dynamics.}
The retention rates ($FRR$) reveal distinct handling trade-offs. Our framework displays an elevated $FRR$ (1.432 on $D_{\text{tex}}^{\text{gt}}$ and 1.129 on $D_{\text{md}}$), partly due to dataset preprocessing where nested LaTeX subfigures are flattened into standalone assets, which our pipeline faithfully reconstructs as individual units. The LLM baseline, conversely, scores near-1.0 on $D_{\text{tex}}^{\text{gt}}$ but drops significantly on $D_{\text{md}}$, suggesting an unconstrained tendency to consolidate assets without verification. Further qualitative analysis is provided in Appendix C.

\textbf{Execution Efficiency.}
Our hybrid pipeline achieves a $5.1\times$ speedup ($208.5\text{s} \rightarrow 40.6\text{s}$) over the E2E LLM. By bounding neural generation to targeted regions, we reduce input and output token consumption by $24.1\%$ and $60.7\%$, respectively, minimizing the production inference footprint.

\subsubsection{Supplementary Track Evaluation}

Beyond layout and structure, we examine textual preservation and referencing accuracy, data is presented in Appendix D1.  

\begin{table*}[!htbp]
  \centering
  \small
  \begin{tabular}{|l|c|c|c|c|c|}
   \hline
    Configuration & CSR$_1$ & Avg Fatal & Failed & FRR & All Tokens \\
    \hline
    Ours & 0.946 $\pm$ 0.227 & 8.7 $\pm$ 6.1 & 3 & 1.432 $\pm$ 0.748 & 25027 $\pm$ 13631 \\
    A1 & 0.911 $\pm$ 0.288 & 7.8 $\pm$ 5.3 & 5 & 1.415 $\pm$ 0.776 & 18772 $\pm$ 9831 \\
    A2 & 0.964 $\pm$ 0.187 & 16.5 $\pm$ 3.5 & 2 & 0.000 $\pm$ 0.000 & 2988 $\pm$ 389 \\
    A3 & 0.929 $\pm$ 0.260 & 12.0 $\pm$ 6.1 & 4 & 1.422 $\pm$ 0.761 & 25006 $\pm$ 13925 \\
    A4 & 0.786 $\pm$ 0.414 & 5.8 $\pm$ 4.3 & 12 & 1.405 $\pm$ 0.775 & 21557 $\pm$ 13672 \\
    A5 & 0.946 $\pm$ 0.227 & 7.7 $\pm$ 6.0 & 3 & 1.436 $\pm$ 0.751 & 24824 $\pm$ 13292 \\
    A6 & 0.893 $\pm$ 0.312 & 8.2 $\pm$ 5.6 & 6 & 1.393 $\pm$ 0.785 & 38785 $\pm$ 16686 \\
    \hline
  \end{tabular}
  \caption{Compilation robustness under varied ablation configurations.}
  \label{tab:ablation_results}
\end{table*}

\textbf{Semantic Similarity.} Across all venues, both theLLM baseline and our framework yield consistently high and closely aligned scores in both ROUGE-L \citep{Lin:2004jd} and BERTScore F1 \citep{Zhang:2019jd} metrics, demonstrating that both paradigms successfully preserve the underlying paper content without introducing perceptible semantic drift.

\textbf{Citation Integrity.} Our framework achieves a citation form accuracy of $0.995$, exceeding the LLM baseline and remains stable across different venues. While the LLM baseline maintains a slightly higher citation recall and exhibits larger variance across different citation formats. This suggests that deterministic structural handlers, by enforcing stricter matching criteria to achieve higher precision, offer higher operational stability for structured academic references. Detailed cross-venue citation distributions and explanations are provided in Appendix D.

\subsection{Ablation Study}
To evaluate the individual contributions of core components within our hybrid pipeline, we conduct six targeted ablation experiments across the 56-document corpus: A1 replaces venue-specific manifest $\mathcal{M}(\mathcal{T})$ with generic text guidelines; A2 strips asset placeholders implemented during Stage $T_2$ prior to assembly; A3 downgrades the LLM model in the hybrid pipeline from Qwen-Plus \cite{yang2024qwen2technicalreport} to the lower-parameter Qwen-Flash \citep{yang2024qwen2technicalreport}; A4 replaces the neural global document assembly layer with rule-based string concatenation (Stage $T_5$); A5 bypasses the deterministic inline validation layer (Stage $T_4$) appended to $T_3$; and A6 feed all chunks through the LLM instead of asset chunks only. Table \ref{tab:ablation_results} summarizes the aggregate impacts of these architectural shifts on compilation robustness, with extended metrics detailed in Appendix E.

\textbf{Empirical Insights.} The quantitative results uncover three primary engineering trade-offs. First, results highlights the necessity of neural assembly layers; A4 drops $CSR_1$ to $0.786$, as rigid rules fail to assemble body text to preambles and references without conflicts. Second, A2 creates an oversimplified optimization illusion, inflating $CSR_1$ to $0.964$ but doubling average fatal errors ($\bar{e} = 16.50$) and losing all assets, indicating that the placeholder design is vital for asset preservation. Third, A5 yields identical aggregate success rates, showing that while it serves as a safeguard against structural breaks, its marginal utility for syntax repair diminishes when deploying highly capable base models. Finally, A6 reduces stability ($CSR_1 = 0.893$) and increases operational costs. This penalty demonstrates the utility of our selective, asset-targeted allocation framework.

%% file: 05_Discussion.tex
\section{Discussion}
The evaluation of our framework yields critical insights for production-grade document conversion. First, the boundary between deterministic syntax mapping and probabilistic neural inference directly determines system efficiency. Second, long-context prompts may trigger structural drift and error propagation that require exhaustive manual audits. Our architectural decoupling enables localized verification allowing targeted debugging that monolithic pipelines lack. Finally, monolithic inputs are susceptible to API safety-filter disruptions. By segmenting documents, our framework restricts these triggers to localized snippets or bypasses them entirely, preventing systemic failures and ensuring continuous service availability. A comprehensive analysis of these failure modes and architectural trade-offs is provided in Appendix G.

%% file: 06_Conclusion.tex
\section{Conclusion}
This work resolves the trade-offs between rule-based converters and monolithic end-to-end LLM generation in template-constrained document transformation. We introduce the \textbf{Dual-Track Framework}, which systematically decouples structural template compliance from semantic translation. By executing a hybrid pipeline, the framework confines LLM inference strictly to reasoning-intensive segments while routing deterministic syntax to rule-based engines. Evaluation across 56 published papers demonstrates that this approach substantially improves compilation success rates with higher time and cost efficiency under production-like conditions. Looking ahead, we plan to extend this Dual-Track framework to other structured formatting standards such as HTML/XML, providing an auditable, cost-effective blueprint for enterprise document automation.

%% file: 07_Limitations.tex
\section{Limitations}

\textbf{Asset translation rigidity.} The chunk translator (T3) currently operates on individual figure and table placeholders,  lacking the ability to handle multi-panel sub-figures or custom layouts. Additionally, the conservative mapping strategy occasionally causes asset duplication across chunk boundaries, requiring manual cleanup. Context-sensitive syntax selection and asset deduplication remain future work.

\textbf{Robustness under High Complexity.} The $T_3$–$T_4$ pipeline has a finite robustness envelope. Extremely high asset density (more than 10 assets clustered together) can cause the LLM to drop placeholder references, while highly complex tables (e.g., extra-wide columns with non-standard markup or non-English Unicode) can lead to parsing failures during the checking stage.

\textbf{Limited evaluation scope.} Our benchmark is currently restricted to English-language papers in computer science and artificial intelligence. The generalization our framework to other academic disciplines, non-English languages, and diverse citation or layout styles remains untested.

%% file: 08_Ethical_Considerations.tex
\section{Ethical Considerations}

\textbf{Data acquisition.} Our benchmark is constructed from papers retrieved from the arXiv public search API. All source documents are publicly available under the open-access terms of their respective arXiv licenses. We use this data solely for academic evaluation purposes: no derivative dataset is commercialized, redistributed, or publicly released..

\textbf{Assistive tool, not a content generator.} Our framework performs format conversion rather than content generation. The input is authored by the human researcher; the system produces no novel scientific claims, rewrites no arguments, and contributes no intellectual content beyond typesetting decisions. This cleanly separates our work from end-to-end LLM baselines that generate complete documents from minimal prompts: our sparse-allocation architecture confines the LLM to structural rendering subtasks (figure and table placement, bibliography formatting), while the semantic core of the paper remains exclusively the author's responsibility.

\textbf{Deployment and accountability.} When embedded in a publisher's submission or production pipeline, the system acts as an assistive typesetting tool with a human in-the-loop. The final responsibility for verifying the correctness of the compiled output, including mathematical fidelity, citation accuracy, and template compliance, rests with the system \textit{user} (the author or editorial staff), not the system provider. We recommend that every automatically generated \texttt{main.tex} undergo human review before submission or publication, particularly in venues where formatting errors may affect review outcomes.

%% file: 10_Appendix_A.tex
\section{Mathematical Formalization of the Dual-Track Mapping}

Let $\mathcal{T}$ be the target LaTeX template and $D_{\text{md}}$ be the source Markdown document modeled as an ordered sequence of disjoint content elements, $D_{\text{md}} = (t_1, t_2, \dots, t_n)$, with a coordinate index set $I = \{1, 2, \dots, n\}$.

The global conversion mapping is formally decomposed into an offline profile extraction $f_{\text{off}}$ and an online execution pipeline $f_{\text{on}}$:
\[\mathcal{M}(\mathcal{T}) = f_{\text{off}}(\mathcal{T}), \quad \text{and} \quad D_{\text{tex}} = f_{\text{on}}(D_{\text{md}} \mid \mathcal{M}(\mathcal{T})).\]

To execute $f_{\text{on}}$, the index set $I$ is partitioned into three pairwise disjoint functional regions based on our task region taxonomy: $I = I_{\text{sym}} \cup I_{\text{prob}} \cup I_{\text{ns}}$. The online conversion mapping $f_{\text{on}}$ is thus evaluated via localized operators tailored to each specific coordinates:

\[f_{\text{on}}(t_i \mid \mathcal{M}(\mathcal{T})) = \begin{cases} g_{\text{rule}}(t_i), & i \in I_{\text{sym}} \\ h_{\text{llm}}(t_i \mid \mathcal{M}(\mathcal{T})), & i \in I_{\text{prob}} \\ v_{\text{rule}}(h_{\text{llm}}(t_i \mid \mathcal{M}(\mathcal{T}))), & i \in I_{\text{ns}} \end{cases},\]

where $g_{\text{rule}}$ represents deterministic rule-based engines, $h_{\text{llm}}$ denotes LLM inference, and $v_{\text{rule}}$ signifies the deterministic verification layer enforcing syntactic invariants. The final LaTeX artifact is assembled via sequence concatenation ($\bigoplus$):
\[D_{\text{tex}} = \bigoplus_{i=1}^{n} f_{\text{on}}(t_i \mid \mathcal{M}(\mathcal{T})).\]

%% file: 11_Appendix_B.tex
\section{Evaluation Dataset Characteristics and Preparation Pipeline}

\subsection{Dataset Diversity and Template Distribution}

The evaluation dataset comprises 56 published, camera-ready papers sampled across 7 distinct publishing venues which represents an evaluation scale totaling over 500 pages of dense academic content to ensure layout diversity and prevent single-template overfitting. The specific template distributions and their structural characteristics are detailed below:

\begin{itemize}
  \item \textbf{Computer Science and AI Venues:} 48 papers are evenly sampled from AAAI, ACL, CVPR, ICML, ICLR, and NeurIPS (8 papers per venue, excluding surveys and workshop papers). This subset covers a balanced mix of single-column formats and dense double-column layouts bound by strict page-budget constraints.
  \item \textbf{IEEE Multi-Domain Subset:} 8 papers are sampled from the IEEE template family to evaluate the framework's cross-domain generalizability. To test beyond standard AI layouts, the selection spans both flagship robotics/systems conferences—the IEEE International Conference on Human-Robot Interaction (\textbf{HRI}) and the IEEE International Conference on Intelligent Transportation Systems (\textbf{ITSC})—and specialized engineering domains, specifically the International Conference on Axiomatic Design (\textbf{ICAD/ICAHS}).
\end{itemize}

This composition forces the evaluation to process highly divergent macro packages, distinctive bibliography styles, and contrasting visual/tabular floats, establishing a robust testing environment for production-grade document automation.

\subsection{Ground-Truth Preparation Pipeline}

The dual-ground-truth pairs used in our experiments are generated from reverse-engineering raw camera-ready LaTeX source packages with a deterministic programmatic pipeline consisting of three stages:

\begin{enumerate}
  \item \textbf{LaTeX Ground-Truth Formulation ($D_{\text{tex}}^{\text{gt}}$):} For each selected paper, the multi-file source package (including class files, style configurations, and graphics) is flattened into a single, cohesive root LaTeX file. This file serves as the gold standard for rigid layout compliance and template-specific syntactic invariants.
  \item \textbf{Noise Cleansing and Human-Draft Emulation:} To mimic the syntax density of a human-authored draft while stripping machine-generated artifacts, the raw Markdown stream undergoes aggressive programmatic sanitization. We programmatically strip Pandoc-specific internal noise (including \texttt{[ ]\{\#\}} spans, \texttt{::: \{\}} divs, and grid tables), unresolved common Unicode anomalies, and residual malformed HTML fragments.
  \item \textbf{Target Serialization (\texttt{main.md}):} The resulting clean file, defined as the \textit{Markdown Ground Truth}, explicitly preserves the logical structural hierarchy of the original publication: title, abstract, document body (embedding complex inline/display mathematics, multi-column tabular configurations, nested figures, and raw author-year citation keys), and appendices. Crucially, the raw \textit{References} bibliography section is completely omitted from \texttt{main.md} to force downstream translation models to dynamically resolve citations via explicit reasoning mechanisms.
\end{enumerate}

\subsection{Technical Filtering Criteria and Dataset Statistics}

During the ingest phase, any source package exhibiting non-standard local build systems or fractured image assets was excluded. To preserve the mathematical validity of our diagnostic downstream analysis, a proactive verification layer checks the structural health of each generated asset. Papers exhibiting anomalous layout configurations in their original state—such as a total absence of display equations (\texttt{no\_display\_math}) or missing sectional boundaries—are permanently tagged with metadata \texttt{anomaly\_flags}. These flagged anomalies are isolated and reported separately during error distribution tracking, preventing skewed outliers from contaminating the general performance baseline.

%% file: 12_Appendix_C.tex
\section{Deep-Dive Analysis on Structural and Layout Metrics}

\subsection{Cross-Venue Architectural Distributions}

While Table 2 in the main text summarizes the aggregate performance across the entire 56-document corpus, the specific structural behavior fluctuates across individual publishing templates due to varying macro restrictions.

\begin{itemize}
  \item \textbf{Mathematical Environment Over-Generation:} In monolithic E2E LLM configurations, unconstrained context generation induces severe token-level hallucination within mathematical blocks. This layout drift scales significantly in single-column layouts such as ICLR, where the baseline's mathematical retention rate against the Markdown anchor ($\text{MathER}_{\text{md}}$) inflates to $1.83$. In contrast, our hybrid routing restricts mathematical processing to deterministic synchronization, tracking the exact baseline precision of rule-based engines across all seven venues.
  \item \textbf{Macro-Sectioning Invariants:} Sectioning retention rates ($\text{SRR}$ and $\text{SRR}_{\text{md}}$) remain close to $1.00$ across all tested pipelines. Minor localized variances primarily stem from structural redundancies, such as the duplication of abstract or reference block headers during monolithic generative merges, rather than systemic content omission.
\end{itemize}

\begin{figure*}[!htbp]
  \includegraphics[width=\linewidth]{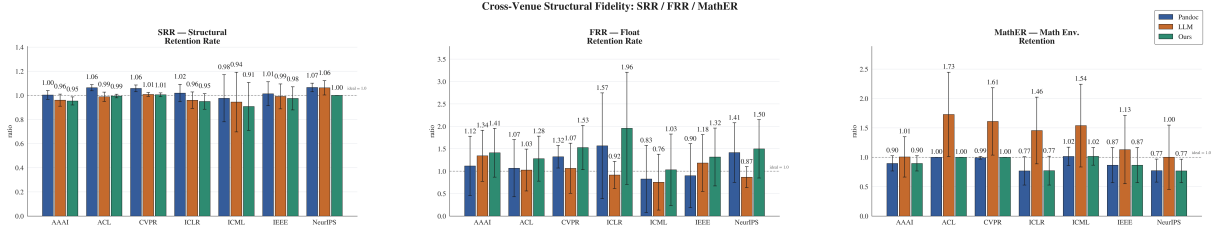}
  \caption{Bar Chart illustrating cross venue fidelity to the LaTeX Ground Truth ($D_{\text{tex}}^{\text{gt}}$)}
  \label{fig:cross_venue_fidelity_tex}
\end{figure*}

\begin{figure*}[!htbp]
  \includegraphics[width=\linewidth]{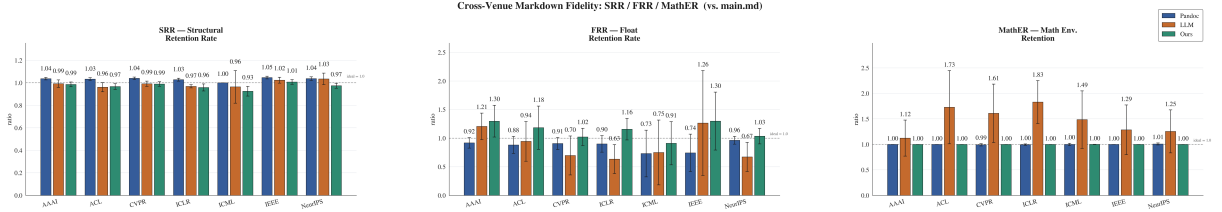}
  \caption{Bar Chart illustrating cross venue fidelity to the Markdown Ground Truth ($D_{\text{md}}$)}
  \label{fig:cross_venue_fidelity_md}
\end{figure*}

\begin{figure*}[!htbp]
  \includegraphics[width=\linewidth]{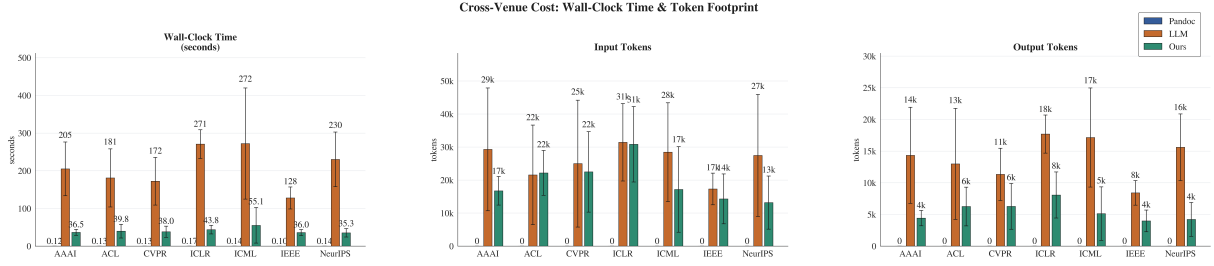}
  \caption{Bar Chart illustrating cross\_venue\_cost}
  \label{fig:cross_venue_cost}
\end{figure*}

\subsection{Float Asset Degradation and Preprocessing Artifacts}

The variance observed in float environment retention rates ($\text{FRR} = 1.432$ for our framework) is driven by two distinct engineering factors:

\begin{enumerate}
  \item \textbf{Upstream Subfigure Flattening:} During the dataset formulation layer, the initial rule-based parsing via Pandoc programmatically flattens nested LaTeX subfigures into individual, standalone asset blocks within the initial Markdown stream. Our online track subsequently processes and outputs these items as standalone figures, resulting in an inflated environment count against the original multi-panel LaTeX ground truth ($D_{\text{tex}}^{\text{gt}}$).
  \item \textbf{Conservative Constraint Overlap:} Within the localized asset handlers ($T_4$), prompt instructions are weighted heavily toward asset preservation to prevent data loss. Near chunk boundaries, this safety constraint occasionally generates duplicate figure environment blocks for identical image tokens. Conversely, the E2E LLM baseline maintains a near-$1.0$ ratio against $D_{\text{tex}}^{\text{gt}}$ but drops underperforms on $D_{\text{md}}$, illustrating an implicit behavior where monolithic pipelines autonomously consolidate adjacent image blocks without structural verification.
\end{enumerate}

\subsection{Cross-Venue Resource and Infrastructure Costs}

The runtime execution overhead represents a primary barrier for production deployment. The rule-based Pandoc engine operates with minimal latency and zero token costs but fails to satisfy template constraints. When comparing neural architectures, the E2E LLM baseline requires processing the entire document context simultaneously, which scales quadratically and causes severe latency degradation (averaging $208.5$ seconds per document). By isolating generative tasks to independent text chunks, our parallel execution layer splits dense token strings, dropping the average processing time to $40.6$ seconds while yielding a $24.1\%$ reduction in input tokens and a $60.7\%$ reduction in generated output tokens.

%% file: 13_Appendix_D.tex
\section{Extended Analysis on Semantic and Citation Metrics}

\subsection{Semantic Preservation}

\begin{figure*}[!htbp]
  \includegraphics[width=\linewidth]{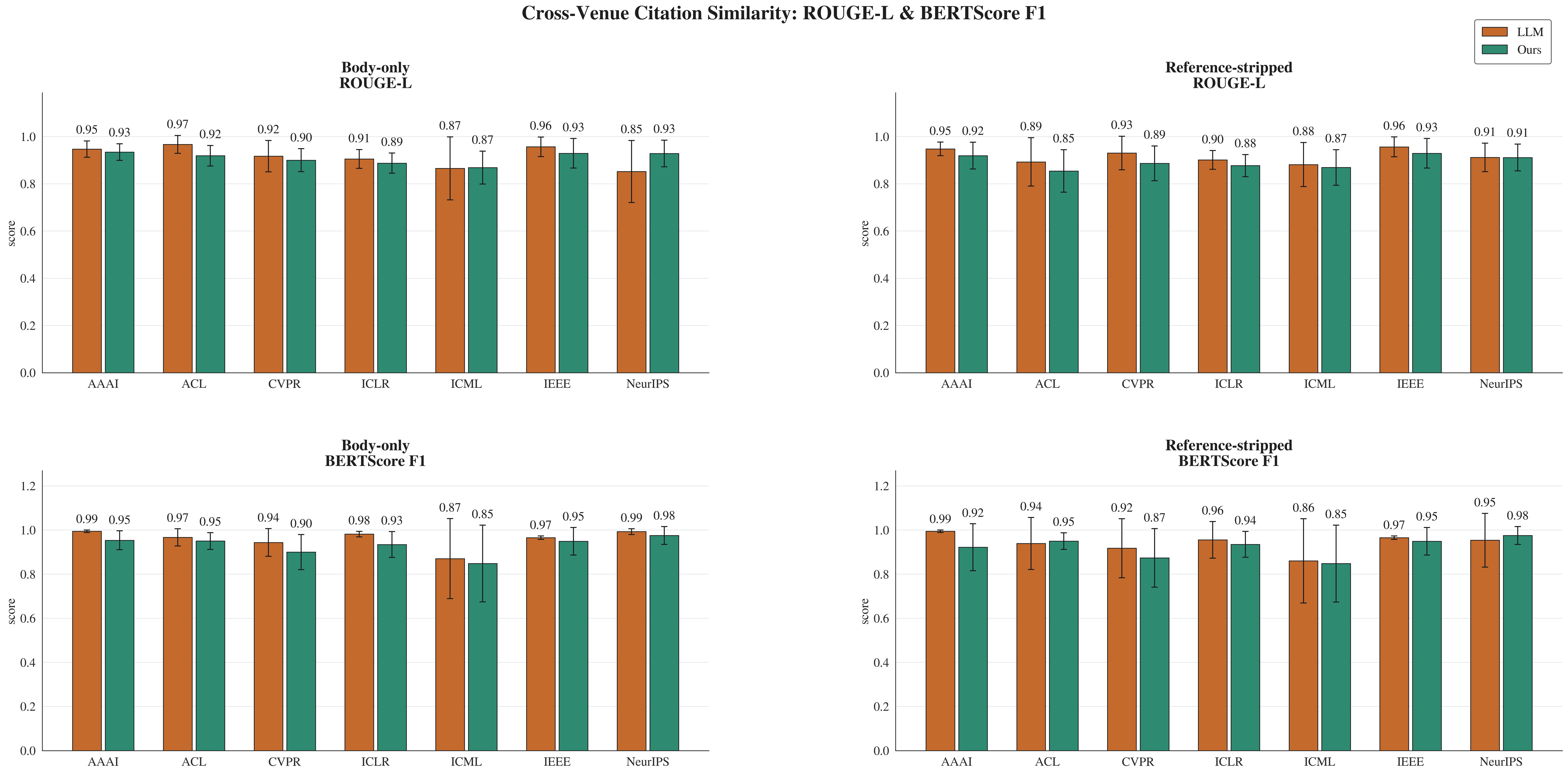}
  \caption{ROUGE-L \citep{Lin:2004jd} and BERTScore F1 \citep{Zhang:2019jd} distribution plots for LLM baseline and our framework.}
  \label{fig:D1}
\end{figure*}

\begin{table*}[!htbp]
\makebox[\textwidth][c]{%
  \centering
  \small
  \begin{tabular}{|l|c|c|c|c|c|c|c|}
  \hline
    Method  & ROUGE-L & BERTScore F1 & ROUGE-L (ref-stripped) & BERTScore (ref-stripped) & Cit. Form Accuracy & Cit. Recall \\
    \hline
    LLM  
        & \makecell{0.916 \\ \small $\pm$ 0.087} 
        & \makecell{0.959 \\ \small $\pm$ 0.081} 
        & \makecell{0.918 \\ \small $\pm$ 0.069} 
        & \makecell{0.941 \\ \small $\pm$ 0.114} 
        & \makecell{0.880 \\ \small $\pm$ 0.279} 
        & \makecell{0.956 \\ \small $\pm$ 0.071} \\\
        hline
    Ours  
        & \makecell{0.910 \\ \small $\pm$ 0.055} 
        & \makecell{0.930 \\ \small $\pm$ 0.088} 
        & \makecell{0.893 \\ \small $\pm$ 0.069} 
        & \makecell{0.922 \\ \small $\pm$ 0.103} 
        & \makecell{\textbf{0.995} \\ \small $\pm$ 0.022} 
        & \makecell{0.919 \\ \small $\pm$ 0.092} \\
        \hline
  \end{tabular}}
  \caption{Semantic and citation accuracy metrics. \emph{Ref-stripped} denotes metrics evaluated on content excluding bibliography sections.}
  \label{tab:supplementary_metrics}
\end{table*}

ROUGE-L \citep{Lin:2004jd} and BERTScore F1 \citep{Zhang:2019jd} distributions indicate that both the monolithic LLM and our hybrid framework achieve high content overlap with the ground truth. The minor variances observed in the BERTScore (body) reflect localized phrasing differences rather than systemic content omission. Detailed distribution plots for each of the 7 venues are deferred to Figure \ref{fig:D1}.

\subsection{Citation Performance Trade-offs}

The performance gap in citation metrics stems from a structural trade-off:
\begin{figure*}[!htbp]
  \includegraphics[width=\linewidth]{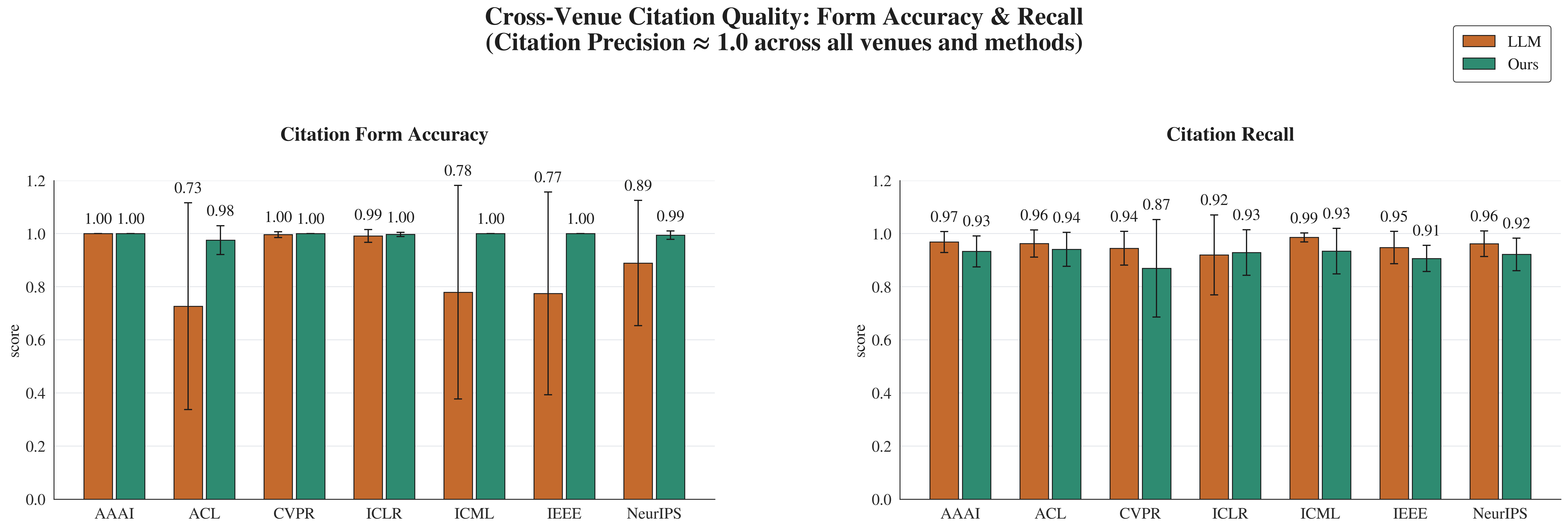}
  \caption{Citation Form Accuracy and Recall distribution plots for LLM baseline and our framework.}
  \label{fig:cross_venue_citation_quality}
\end{figure*}

\begin{itemize}
  \item \textbf{Syntactic Precision:} Our framework’s high form accuracy ($0.995$) results from deterministic BibTeX parsing, which enforces template-specific citation commands.
  \item \textbf{Recall Sensitivity:} The baseline's higher recall ($0.956$) is attributed to its monolithic nature, allowing it to recognize diverse in-text citation formats and infer citation labels when original BibTeX keys are ambiguous. Our framework’s lower recall ($0.919$) arises from its strict handling of ambiguous citation strings—where the system prioritizes structural validity and prevents invalid macro generation at the cost of excluding malformed or unrecognizable citation items from the final output.  This trade-off yields higher system robustness, traceability, and transparency; furthermore, any unmatched citations are systematically recorded in an extraction log file for subsequent manual review.
\end{itemize}

%% file: 14_Appendix_E.tex
\section{Modular Analysis and Extended Ablation Metrics}

\begin{figure*}[!htbp]
  \includegraphics[width=\linewidth]{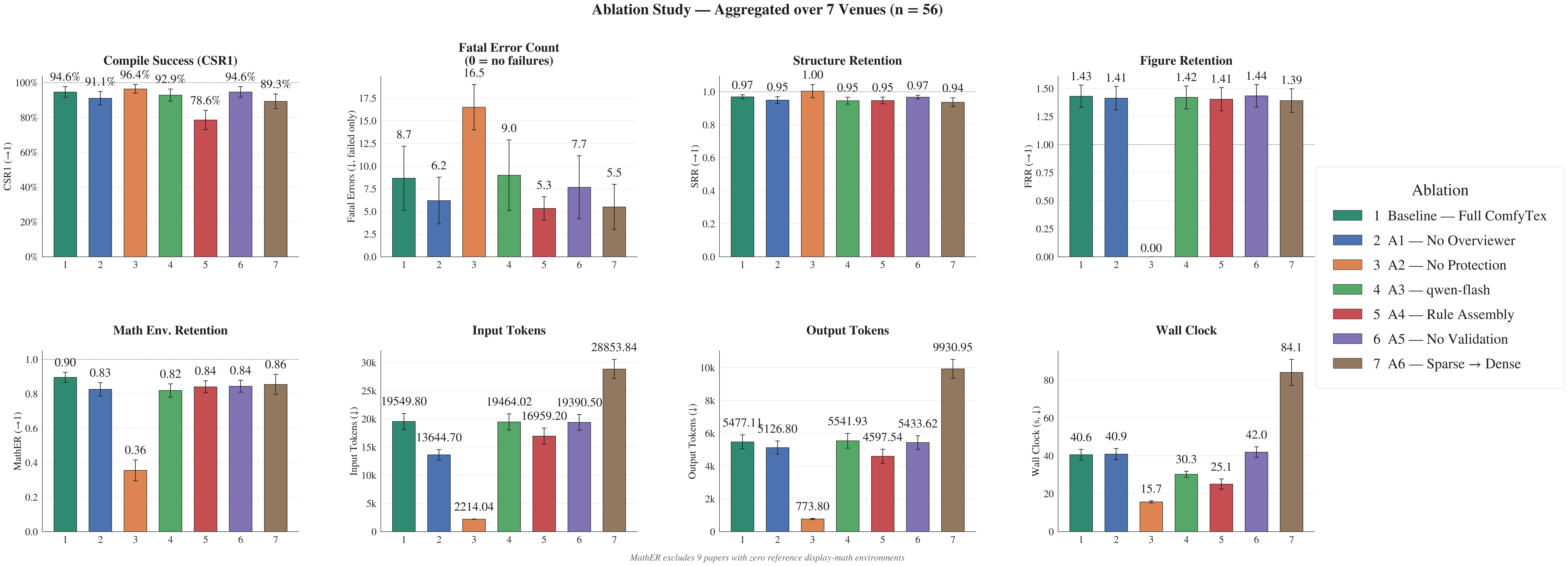}
  \caption{Comprehensive modular breakdowns across eight discrete metrics, aggregated over 7 publishing venues (n = 56). Columns 1--7 represent the Full System and configurations A1--A6, respectively.}
  \label{fig:ablation_2x4}
\end{figure*}

This appendix provides an extended diagnostic analysis of the six configuration variants detailed in Section 4.3. While Table 4 outlines primary compilability perspective, the multi-panel evaluation illustrated in Figure \ref{fig:ablation_2x4} uncovers the downstream effects of isolating or disabling specific pipeline constraints.

\subsection{The Structure-omission Artifact under Protected Placeholder Removal (A2)}

As quantified in Table 4, removing asset placeholders (A2) presents an anomalous performance signature: the compilation success rate peaks at $0.964 \pm 0.187$ with only 2 failed documents, yet the average fatal error count for those remaining failures nearly doubles to $16.5 \pm 3.5$. An inspection of the underlying metric distributions in Figure \ref{fig:ablation_2x4} explains this behavioral divergence:

\begin{itemize}
  \item \textbf{Asset Environment Erasure:} Stripping the asset placeholders entirely drops the floating environment retention rate to $0.000 \pm 0.000$. Without explicitly protected coordinates, the language model fails to identify the underlying tabular and figure macros during final text rendering.
  \item \textbf{Mathematical and Token Collapse:} This omission pattern extends to mathematical syntax. As tracking indicators, the mathematical environment retention rate ($MathER$) plummets to $0.35$, while input and output tokens collapse sharply to a total footprint of $2,988 \pm 389$ tokens per document ($2,214.04$ input and $773.80$ output tokens in Figure \ref{fig:ablation_2x4}).
  \item \textbf{The Compilability Explanatory Factor:} Consequently, the elevated $CSR_1$ under \textit{A2} does not reflect superior template adaptation. Instead, by translating complex structural layouts into empty text, the model generates trivial, flattened source files that compile effortlessly under local engines at the cost of total content degradation.
\end{itemize}

\subsection{Token-Level Congestion under Full-Chunk Inference (A6)}

The control trajectory of configuration A6, where all segmented text chunks are routed through the neural inference engine instead of confining execution strictly to asset-bearing chunks, reveals severe processing degradation across compute bounds.

\begin{itemize}
  \item \textbf{Inference Footprint Inflation:} Routing deterministic sections through language models inflates the average total token footprint to $38,785 \pm 16,686$ tokens per document ($28,853.84$ input and $9,930.95$ output tokens in Figure \ref{fig:ablation_2x4}). This represents an approximate $55\%$ inflation over the asset-targeted allocation baseline ($25,027 \pm 13,631$ total tokens).
  \item \textbf{Latency Scaling:} This execution burden reflects directly on wall-clock latency, which more than doubles from $40.6$ seconds in the full system setup to $84.1$ seconds under A6.
  \item \textbf{Formatting Drift Analysis:} Because deterministic text strings are subjected to unnecessary open-ended token regeneration, the aggregate compilation success rate drops to $0.893 \pm 0.312$. This confirms that exposing predictable paragraph structures to probabilistic neural mapping introduces formatting drift and localized token variations, validating the system efficiency gained by selective neural isolation.
\end{itemize}

\subsection{Micro-Analysis of Baseline Degradations (A1, A3, A4, A5)}

The remaining modular variations map localized stability components within the hybrid track:

\begin{itemize}
  \item \textbf{Macro Constraints (A1):} Bypassing the venue-specific manifest (\textit{A1}) degrades template compliance ($CSR_1 = 0.911 \pm 0.288$), as the underlying models must infer complex macro packages from generic instruction boundaries.
  \item \textbf{Parameter Capacity Scaling (A3):} Downgrading the foundational model to the lower-parameter \textit{Flash} variant (A3) shifts the failure distribution, increasing the average fatal errors to $12.0 \pm 6.1$, which demonstrates that lower-capacity base models are more susceptible to complex layout corruptions.
  \item \textbf{The Structural Necessity of Model Assembly (A4):} Substituting the neural document assembly layer with rule-based string concatenation (A4) triggers a severe systemic failure mode ($CSR_1 = 0.786 \pm 0.414$, with 12 failed documents). This degradation highlights the necessity of neural tracking layers; when local text blocks are split across parallel workers, static programmatic concatenation cannot dynamically retrieve broken inter-chunk macro states.
  \item \textbf{Online Verification Safeguard (A5):} Removing the online validation layer (A5) maintains an identical success baseline ($0.946 \pm 0.227$) with a minor shift in internal error distributions ($7.7 \pm 6.0$). This confirms that while the layer acts as a proactive safeguard against syntax breaks, frontier models operating on bounded local segments natively output high syntax reliability, leading to diminishing marginal returns for automated downstream code repair.
\end{itemize}

\begin{figure*}[htbp]
  \includegraphics[width=\linewidth]{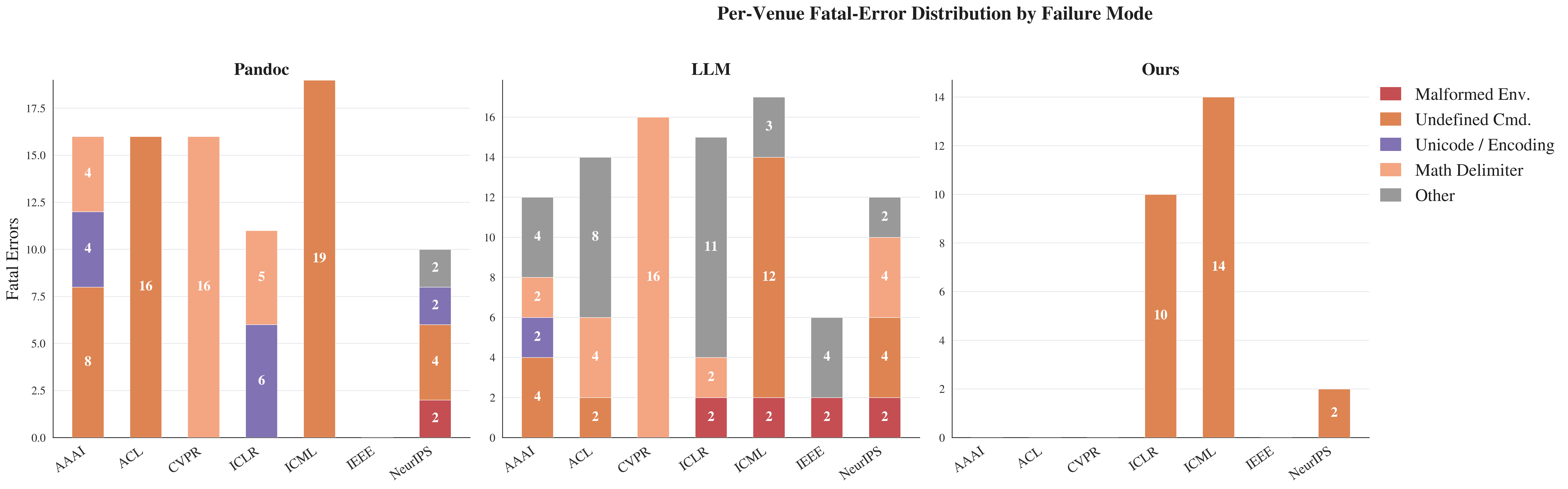}
  \caption{Detailed Failure mode and distribution per venue}
  \label{fig:f2}
\end{figure*}

%% file: 15_Appendix_F.tex
\section{Failure Analysis}
To gain deeper insights into the results, we conduct an analysis of compilation failures across our 56-paper benchmark. As summarized in the failure mode distributions in Figures \ref{fig:f1} and \ref{fig:f2}, our framework significantly reduces the total number of fatal compilation errors compared to both the rule-based Pandoc engine and the monolithic E2E LLM baseline.

\subsubsection{Shared Limitations and Preprocessing Constraints}
Across the entire corpus, three papers failed to compile under all three pipelines. The root cause for these failures originates from our dataset preprocessing pipeline: during the conversion of original source packages into Markdown Ground Truth ($D_{\text{md}}$), the flattening process retained complex, venue-specific custom macros (e.g., \texttt{\textbackslash{}ERLag}, \texttt{\textbackslash{}ALoss}) within the body text. As these macros are not explicitly defined in the standard venue templates, all three pipelines regardless of their conversion strategy encountered fatal undefined control sequence errors. We identify this as a limitation of our current preprocessing pipeline, as algorithmically identifying and sanitizing all user-defined macros without compromising the structural integrity of the ground truth remains an open reproducibility challenge. Consequently, these three papers contribute to the failure count for our system, including the fatal error outlier observed in the ICML venue (see Figure \ref{fig:f2}).

\subsubsection{Failure Modes: LLM Baseline}

\begin{figure*}
    \includegraphics[width=\linewidth]{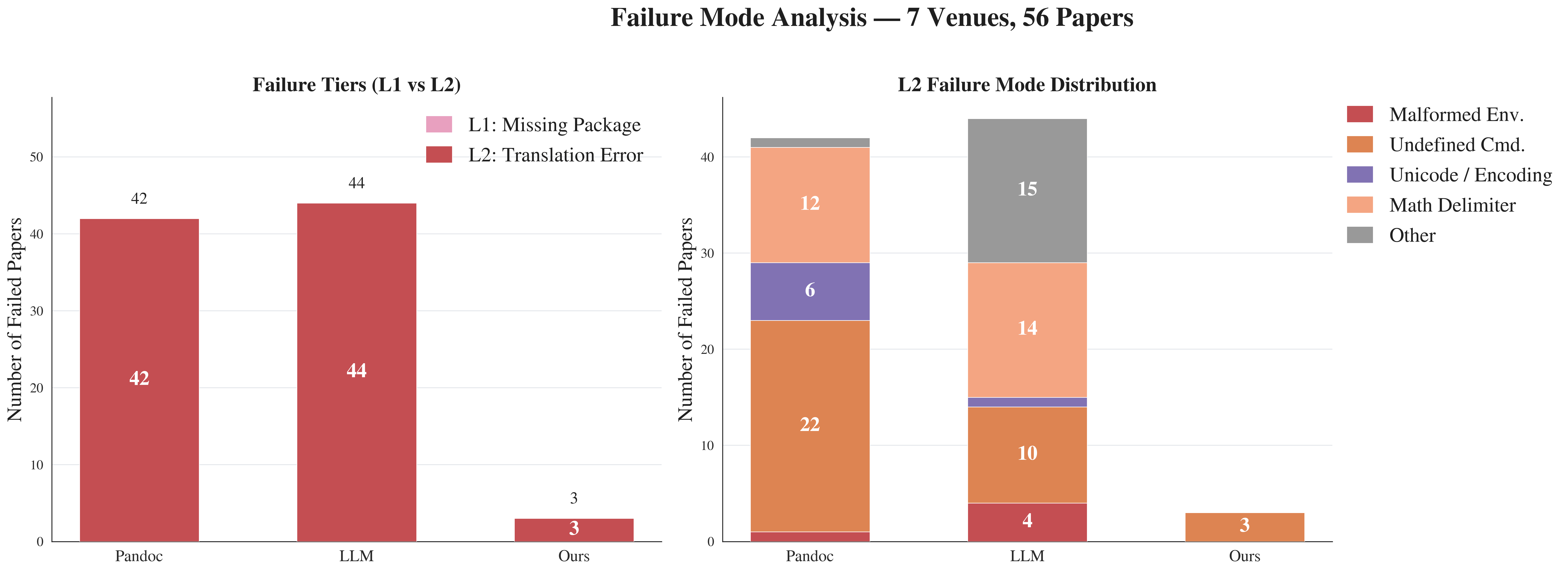}
  \caption{Overall failure modes distribution of all pipelines}
  \label{fig:f1}
\end{figure*}

The monolithic LLM baseline exhibits two primary categories of failure:

\begin{itemize}
  \item \textbf{Translation-Induced Defects:} Notably, all CVPR instances failed under the LLM baseline. Investigation revealed that the standard CVPR template’s \texttt{\textbackslash{}maketitle} command triggers an inherent math-mode error when compiled with the \texttt{-halt-on-error} flag. While our evaluation harness explicitly disables this flag for the Pandoc pipeline to ensure fairness, the LLM baseline, by autonomously generating its own compilation command, inadvertently retained this flag, leading to premature termination. This highlights a critical limitation in monolithic LLM-based pipelines, which often struggle to reason about template-specific compilation nuances.
  \item \textbf{Generative Hallucinations:} Beyond template-specific issues, the LLM baseline incurred failures across 8 unique instances due to hallucinated syntax, including \texttt{missing\_math\_delimiter}, \texttt{undefined\_control\_sequence},  \texttt{malformed\_environment} and \texttt{other} errors. These failures reinforce our hypothesis that delegating deterministic mapping to purely probabilistic neural models leads to structural drift and fragile output.
\end{itemize}

\subsubsection{Failure Modes: Pandoc Baseline}

Pandoc’s failures were primarily confined to \texttt{unicode\_error} and \texttt{undefined\_control\_sequence} issues. These instances typically occurred in documents containing non-Latin characters or legacy bibliographic command structures that the rule-based engine failed to map correctly. In contrast, our Dual-Track Framework effectively mitigates these failure modes. By routing standard text through deterministic mapping and utilizing the inline structural validation layer (Stage $T_4$) for neural-generated snippets, our system successfully resolves the majority of \texttt{unicode\_error} and basic syntax deviations that frequently derail the Pandoc and LLM baselines. This robustness across diverse venues, as illustrated in the failure mode distribution (Figure \ref{fig:f1}), demonstrates the efficacy of our neuro-symbolic approach in document automation.

%% file: 16_Appendix_G.tex
\section{Extended Discussion on Architectural Trade-offs}

\subsection{The Symbolic-Neural Boundary Principle}

The boundary between deterministic and probabilistic routing directly determines runtime efficiency and compilation success. Rule-based parsing handles structured, invariant syntax with provable correctness, whereas neural inference is reserved strictly for open-ended or multimodal reasoning. Utilizing LLMs for deterministically solvable tasks increases computational complexity and introduces unnecessary failure modes. Confining neural execution exclusively to genuinely underdetermined segments optimizes both compilation robustness and operational resource footprints.

\subsection{Prompt Fragility and Context-Scale Degradation}

Pure prompt-based constraints suffer from performance degradation as the input context scales. Within short local text windows, generative models robustly adhere to formatting instructions, making inline validation layers a passive fallback. However, expanded context windows and dense token dependencies induce structural and layout hallucinations \citep{Liu:2024jd}. This causes monolithic end-to-end LLM pipelines to fail when handling complex venue templates, leading to severe compilation instability.

Furthermore, monolithic LLMs suffer from silent error propagation, as standard prompts cannot programmatically verify downstream instruction compliance to intercept structural drifts (e.g., resulting in inflated mathematical environment counts due to unclosed tags). Auditing these errors in dense token blocks requires manual code reviews. By decoupling document segments and logging discrete execution errors, our framework replaces global human inspection with deterministic tracking and targeted, automated debugging.

\subsection{Content Truncation and Alignment Safety}

Monolithic LLM pipelines face severe engineering vulnerabilities from third-party content alignment and safety policies. During benchmarking, full-document generative requests sometimes causes job failures when harmless, domain-specific scientific terminology (e.g., specific medical classifications or network security nomenclature) inadvertently triggers provider-side API safety filters.

Our framework's parallel chunking architecture explicitly mitigates this systemic risk by isolating individual content segments. When a false-positive safety trigger occurs, it remains strictly localized to a single text snippet, allowing the rest of the document structure to compile normally. For enterprise-grade workflows, this decentralized execution prevents unexpected document-wide data loss and minimizes system downtime.